\documentclass[10pt,twocolumn,letterpaper]{article}

\usepackage{cvpr}
\usepackage{times}
\usepackage{epsfig}
\usepackage{graphicx}
\usepackage{amsmath}
\usepackage{amssymb}
\usepackage{hyperref}
\usepackage[flushleft]{threeparttable}
\usepackage{booktabs} 
\usepackage{amsfonts}
\usepackage{multirow}
\usepackage{tabularx}
\usepackage{xcolor}
\usepackage{footnote}
\usepackage{caption}
\usepackage{cleveref}
\usepackage{enumitem}
\usepackage{wrapfig}
\usepackage{url}
\usepackage{array}
\usepackage[T1]{fontenc}
\newcolumntype{?}{!{\vrule width 1pt}}

\newcolumntype{M}{>{$}c<{$}}
\newcolumntype{C}{>{\centering\arraybackslash}p}
\newcolumntype{Y}{>{\centering\arraybackslash}X}

\begin{document}
\title{On Demographic Bias in Fingerprint Recognition}

\author{Akash~Godbole,~Steven~A.~Grosz,~Karthik~Nandakumar~and~Anil~K.~Jain,~\emph{Life~Fellow,~IEEE}
\thanks{A. Godbole, S.A. Grosz, and A.K. Jain are with the Department of Computer Science and Engineering, Michigan State University, East Lansing, MI, 48824 USA (e-mail: godbole1@cse.msu.edu, groszste@cse.msu.edu, jain@cse.msu.edu). Karthik Nandakumar is with Mohamed Bin Zayed University of Artificial Intelligence (MBZUAI), Abu Dhabi, UAE (e-mail: karthik.nandakumar@mbzuai.ac.ae).}
}

\maketitle
\pagestyle{plain}

\begin{abstract}
Fingerprint recognition systems have been deployed globally in numerous applications including personal devices, forensics, law enforcement, banking, and national identity systems. For these systems to be socially acceptable and trustworthy, it is critical that they perform equally well across different demographic groups. In this work, we propose a formal statistical framework to test for the existence of bias (demographic differentials) in fingerprint recognition across four major demographic groups (white male, white female, black male, and black female) for two state-of-the-art (SOTA) fingerprint matchers operating in verification and identification modes. Experiments on two different fingerprint databases (with 15,468 and 1,014 subjects) show that demographic differentials in SOTA fingerprint recognition systems decrease as the matcher accuracy increases and any small bias that may be evident is likely due to certain outlier, low-quality fingerprint images.
\end{abstract}

\begin{figure}[t]
\begin{center}
\includegraphics[width=\linewidth]{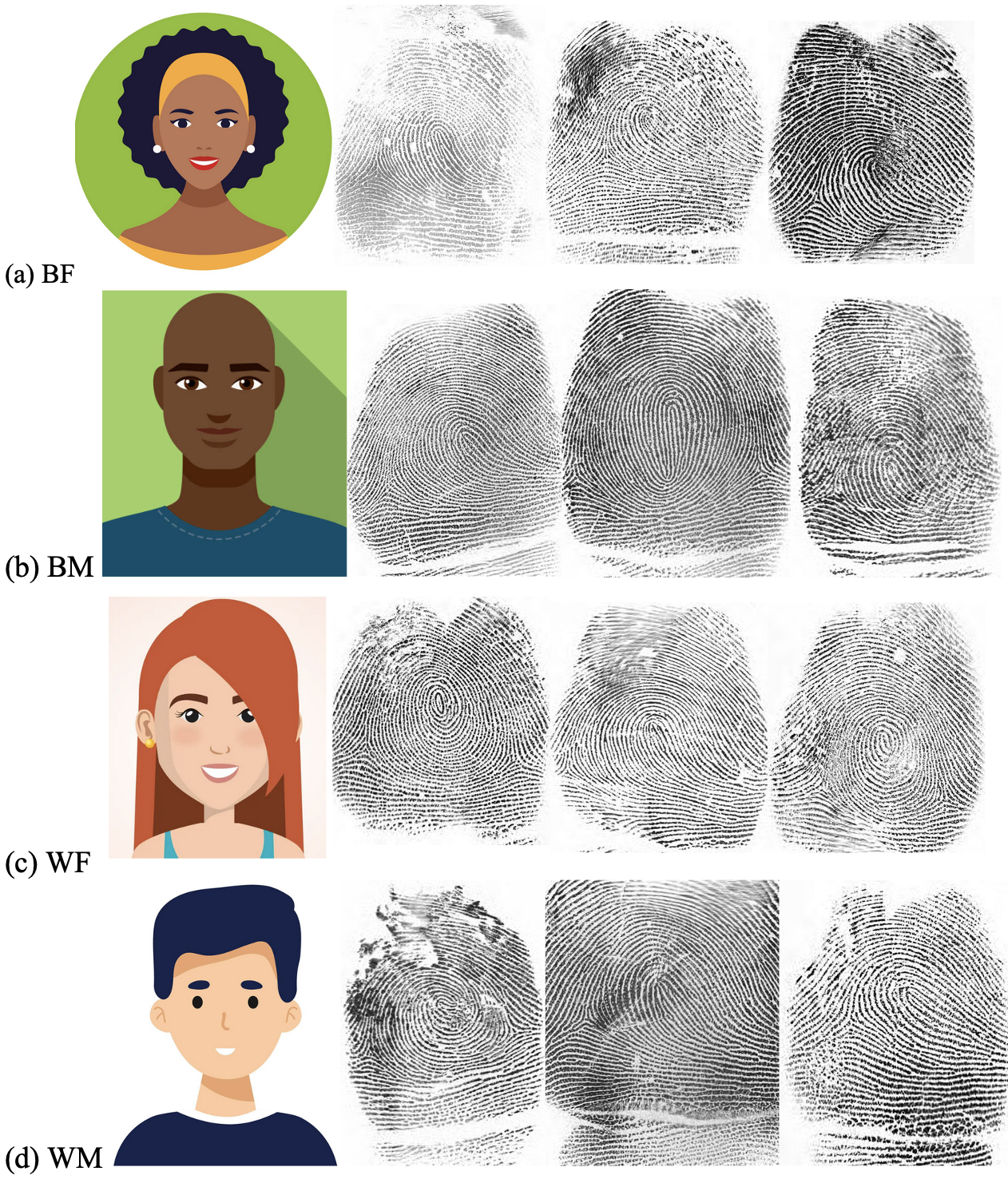}
\captionsetup{justification=centering}
\caption{Sample fingerprint images of four demographic groups examined in this study; (a) Black Female (BF), (b) Black Male (BM), (c) White Female (WF) and (d) White Male (WM). For each group, three fingerprint images are shown along with an animated face to visualize the demographic group. Unlike face images, there are no obvious visual differences between fingerprints of subjects from different demographic groups.}
\label{fig:face_images}
\end{center}
\vspace{-1.5em}
\end{figure}

\section{Introduction}

Fingerprint recognition systems have shown impressive recognition accuracy in recent years \cite{fvc_results}. Consequently, they are now deployed across applications ranging from mobile phone unlock \cite{touchid} to large-scale national identity systems \cite{aadhaar}. However, there is a growing debate about whether biometric systems are fair and equitable. Though there is no clarity yet on the precise definition of fairness, a general consensus is that biometric systems must exhibit similar recognition accuracy across various demographic groups. 

Fairness\footnote{In this study, we use the term fairness in the limited context of absence of demographic bias or existence of demographic parity.} in biometric systems is quantified in terms of \emph{demographic differentials} in the error rates of the system \cite{grother2019face}. According to \cite{transportation_2022}, American airports processed biometric data of over 9 million international travelers in the first six months of 2021. At this scale of operation, even small differentials in the performance across groups may bear significant consequence for individual users as well as organizations that operate such systems. Hence, evaluating the fairness of biometric systems and mitigating any hidden bias is critical for public trust and social acceptance.

\begin{table*}
    \centering
    \captionsetup{justification=centering}
    \caption{Summary of the two databases used in this study\tnote{*}.}
    \begin{threeparttable}
    \begin{tabular}{|c || c | c ? c | c |} 
         \hline
         Demographic Group & \multicolumn{2}{c?}{\textbf{Database $D_1$}} & \multicolumn{2}{c|}{\textbf{Database $D_2$}} \\ [1.0ex]
         \noalign{\hrule height 1.2pt}
         & \# of subjects & approx.\% of database & \# of subjects & approx.\% of database  \\
         \noalign{\hrule height 1.2pt}
         {\includegraphics[scale=0.15]{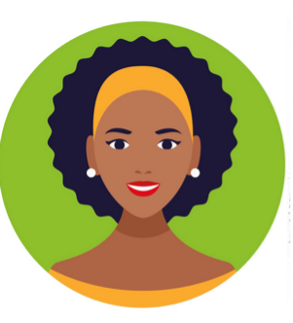}} BF            &  762 & 5 & 172 & 17  \\
         \hline
         {\includegraphics[scale=0.15]{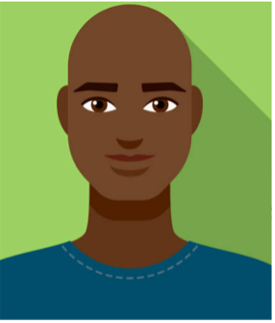}} BM           & 5,594 & 36 & 77 & 8  \\
         \hline
         {\includegraphics[scale=0.15]{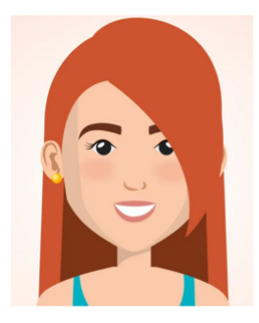}} WF              & 1,195 & 8 & 462 & 45  \\
         \hline
         {\includegraphics[scale=0.15]{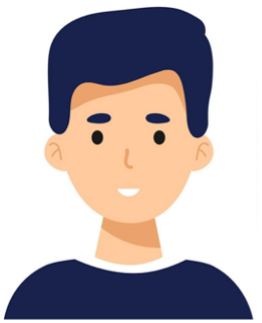}} WM            & 7,917 & 51 & 303 & 30  \\
         \noalign{\hrule height 1.2pt}
    \end{tabular}
    \begin{tablenotes}
    \item[*]{The datasets $D_1$ and $D_2$ are not available in the public domain due to privacy concerns. It is difficult to obtain large-scale fingerprint databases with demographic information in the public domain.}
    \end{tablenotes}
    \end{threeparttable}
    \label{tab:database}
\end{table*}

In the last decade, face recognition systems have come under scrutiny for exhibiting racial and gender bias \cite{najibi_2020, buolamwini2018gender, singer_2019, wingfield_2018, grother2019face}. Several academic publications and media reports have claimed that female and black individuals suffer from lower face detection and recognition accuracies compared to male and white individuals. These claims are a cause of concern due to the proliferation of face recognition systems that have the ability to capture faces at a distance in a covert manner. On the flip side, it has been argued that these criticisms are not justifiable because the claimed differences in face recognition accuracy across demographic groups are not significant. Unlike scientific publications \cite{grother2019face}, commonplace media claims about bias in face recognition often fail to account for rapid progress in facial recognition algorithms in the past 10 years - labelling a system ``racist'' based on evaluations done more than a decade ago \cite{rankonefr, lawfarebias}. 

Leaving aside specific claims and counterclaims about the fairness of facial recognition technology, any evaluation of fairness in biometric systems must be based on state-of-the-art (SOTA) recognition technology and objectively rely on sound analysis as opposed to public sentiment \cite{jain2021biometrics}. Of the existing literature on fairness of fingerprint recognition systems~\cite{hicklin2002implications, sickler2005evaluation, modi2006impact, modi2007impact}, even the most recent study by Yoon et al.~\cite{yoon2015longitudinal} was conducted nearly ten years ago. Given that fingerprint matchers have continued to improve (e.g., the ongoing Fingerprint Verification Competition has seen the Equal Error Rate of fingerprint matchers decrease from 0.1\% to 0.01\% since 2011 \cite{fvc_results}), a fresh look at the fairness of SOTA fingerprint systems becomes an urgent necessity.

Another major lacunae in most of the existing works on fairness in biometric systems is the lack of a standard definition for fairness and a statistical framework to test for fairness. For example, it was reported in \cite{yoon2015longitudinal} that demographic covariates such as race and gender have negligible impact on genuine match scores; however, this does not mean that there will be no difference in the true match rate for different demographic groups. Similarly, if a fingerprint system is evaluated on a dataset and it is observed that the true match rate is $99.5\%$ and $99.2\%$ for males and females, respectively, how can we claim that the observed differential is statistically significant? If we change the dataset by adding or removing some individuals or draw a sample from a different population, will the demographic differential still hold? How large should the dataset be, if we want to claim with $95\%$ confidence that the observed demographic differential is statistically significant? To answer these questions, a formal hypothesis testing framework is essential.

To address the above limitations, we make the following three key contributions in this work:

\begin{enumerate}
    \item  We aim to uncover biases (or lack thereof) in a SOTA commercial-off-the-shelf (COTS) fingerprint matcher (Verifinger 12.3 SDK) and a SOTA deep neural network based fingerprint matcher (DeepPrint \cite{engelsma2019learning}).
    
    \item We propose a statistical testing framework for evaluating bias in biometric systems operating the verification and identification modes.     
    
    \item We apply the proposed statistical framework to test the fairness of SOTA fingerprint systems on two different datasets (see Table \ref{tab:database}) focusing on four demographic groups: black female (BF), black male (BM), white female (WF), and white male (WM). This is aligned with the same demographic groups evaluated in the NIST Face Recognition Vendor Test (FRVT) \cite{grother2019face, grother2022face}. Additionally, the NFIQ2 fingerprint quality metric \cite{nfiq} is computed for the datasets to identify any image quality bias that may exist in the two fingerprint matchers.
\end{enumerate}

\section{Related Work}

\subsection{Definitions of Fairness}

Fairness and demographic bias are well-studied topics in machine learning (ML) systems and several competing notions of fairness have been introduced \cite{MehrabiFairnessMLSurvey2021, liu2021trustworthy}. Typically, a ML system is modeled as a predictor $\hat{Y}$ that predicts an outcome $Y$ based on some input $X \in \mathbb{R}^n$. Ideally, the predictions made by the ML system must not rely on ``protected'' or ``sensitive'' attributes $A$ such as race, gender, and age. For simplicity, let us assume that the true outcome $Y$ (e.g., positive vs negative), the predicted outcome $\hat{Y}$, and the protected attribute $A$ (e.g., white vs black, male vs female) are binary random variables. In this case, the notion of fairness can be formalized in a number of ways.

\noindent \textbf{Statistical Parity}: Also known as \emph{demographic parity}, this criterion requires $P(\hat{Y}|A=0) = P(\hat{Y}|A=1)$. In other words, the predictions are independent of the protected attribute $A$ and the likelihood of observing an outcome should be the same regardless of person's sensitive attribute. However, the above definition of statistical parity is not suitable for biometric recognition systems. For instance, when a male user is attempting to impersonate as a female user in a face recognition system, it is unreasonable to expect the outcome to be independent of the person's gender. 

\noindent \textbf{Equalized Odds}: A predictor $\hat{Y}$ is said to satisfy \emph{equalized odds} if $\hat{Y}$ and $A$ are independent conditional on $Y$ \cite{HardtEqualOpportunity2016}. More specifically, $P(\hat{Y}=1|A=0,Y=y) = P(\hat{Y}=1|A=1,Y =y)$, where $y\in\{0,1\}$. In simple terms, equalized odds ensures equal true positive and false positive rates for the two groups, $A = 0$ and $A = 1$. Since equalized odds allows $\hat{Y}$ to depend on $A$ but through the target $Y$, it is possible to use features that relate directly to $Y$, but not through $A$. This is usually the criterion used in biometric fairness literature, where the difference in the true positive and false positive rates for the two groups is termed as \emph{demographic differential}. Existence of demographic differential indicates that the system violates the equalized odds fairness notion.

\noindent \textbf{Equal Opportunity}: A relaxation to the equalized odds criterion is called \emph{equal opportunity}, which considers only the true positive rates for the two groups and ignores the false positives \cite{HardtEqualOpportunity2016}. Specifically, the system satisfies equal opportunity if $P(\hat{Y}=1|A=0,Y=1) = P(\hat{Y}=1|A=1,Y=1)$. In most biometric applications (especially in the \emph{verification} and \emph{closed set identification} modes), the users want the system to recognize them correctly and the effects of a false match are hardly perceived by the users. Hence, it can be argued that equal opportunity is also an appropriate notion of fairness in biometric systems. 

Apart from the above three widely used definitions of fairness, other fairness criteria have also been proposed in the literature \cite{MehrabiFairnessMLSurvey2021, VermaFairnessDefinitions2018}. Notable among them include \emph{test fairness} (TF), which attempts to calibrate the similarity of a biometric system such that the conditional (genuine and impostor) score distributions are identical across different demographic groups, and \emph{fairness discrepancy rate} (FDR) for biometric verification systems \cite{de2021fairness}, which tries to ensure statistical parity between groups in terms of both false match rate (FMR) and false non-match rate (FNMR) at various decision thresholds using different hyperparameters. Both TF and FDR may be more appropriate during biometric system design, where the designer may not be aware of the operating threshold of the biometric system. In this work, the goal is to evaluate the fairness of operational biometric systems, where the decision threshold has already been fixed. Therefore, equalized odds and equal opportunity are more suitable definitions of fairness.

\subsection{Fairness in Face Recognition}

Studies on fairness of face recognition systems have received a lot of attention due to the high correlation between demographic and facial features. Many of these studies indicate that face recognition systems exhibit some bias against underrepresented ethnic and racial communities \cite{drozdowski2020demographic,terhorst2021comprehensive}. There have been claims and supporting evidence that face recognition has a lower performance for darker skinned people compared to lighter skinned people \cite{najibi_2020}. Another example can be found in \cite{10.2307/26781452}, where it was reported that the classification algorithm built into Google Photos classified some African-American people as ``gorillas". Given such detrimental real-world consequences of demographic differentials in a biometric system (e.g., an innocent person being incarcerated due to a false facial match obtained from surveillance footage), it is important to study the causes for these biases and work towards eliminating them.

\begin{threeparttable}[]
    \centering
    \captionsetup{justification=centering}
    \caption{Demographics of dataset used to train DeepPrint \cite{engelsma2019learning}\tnote{*} }
        \begin{tabular}{|c | c|} 
             \hline
             Demographic Group & \# of unique fingers in training set \\ [1.0ex]
             \hline
             {\includegraphics[scale=0.15]{images/bf.png}} BF            & 1,510    \\
             \hline
             {\includegraphics[scale=0.15]{images/bm.png}} BM              & 17,090      \\
             \hline
             {\includegraphics[scale=0.15]{images/wf.png}} WF              & 1,750      \\
             \hline
             {\includegraphics[scale=0.15]{images/wm.png}} WM              & 15,760      \\
             \hline
        \end{tabular}
        \begin{tablenotes}
            \item[*] {The total number of unique fingers used to train DeepPrint was 38,291, which also includes a very small number of identities from other races as well as those with undisclosed demographic attributes. \\\\}
        \end{tablenotes}
    \label{tab:dp_train}
\vspace{-1.5em}
\end{threeparttable}

\subsection{Fairness in Fingerprint Recognition}

\par A limited number of studies are available on the fairness of fingerprint recognition \cite{sickler2005evaluation, o2011impact, yoon2015longitudinal, modi2006impact, modi2007impact, hicklin2002implications}.  Marasco \cite{9186012} analyzed biases in fingerprint recognition systems including biases induced by sensors and fingerprint image quality. Though the number of low quality (estimated using NFIQ2 measures) fingerprint images was much higher for females compared to males, it was observed that the recognition accuracy was higher for females. Jain et al. \cite{jain2021biometrics} pointed out that bias in the demographic distribution of the dataset can produce biases in fingerprint recognition systems - especially due to the fact that most modern biometric systems are based on deep neural network (DNN) models. These DNN models generally tend to be prejudiced against the underrepresented demographic groups. For instance, a DNN trained on a dataset that is over sampled with white subjects can be expected to show higher performance on white subjects (see table \ref{tab:dp_train}). This phenomenon has been demonstrated in facial recognition and is referred to as the ``other-race-effect''. This term is generally used when referring to the higher ability of humans to recognize faces of people from their own race \cite{furl2002face}. In terms of fingerprint recognition, the other-race-effect would translate to a positive correlation between the verification performance of the system and the majority demographic group in the training dataset \cite{furl2002face, phillips2011other}.

\subsection{A Note on Vocabulary}

While studying demographic bias, it is important to use precise terminology for describing various demographic traits. The primary demographic attributes considered in this study are race and gender. Present day societal norm is to use the words `race' and `ethnicity' interchangeably. However, `race' is associated with physical attributes, while `ethnicity' relates to an individual's cultural upbringing \cite{blakemore_2021}. Since this study is concerned with the the physical traits of the subjects, the relevant term is \emph{race}. A similar dilemma exists when it comes to the terms `gender' and `sex' \cite{sexgender05}. `Gender' is more appropriate when discussing the self-identity of an individual and how they project themselves to society. On the other hand, `sex' refers to the biological traits present in an individual that classify them as either male or female. Since the subjects in the datasets used in this study presented their own identity, the appropriate term to use is \emph{gender}. 

\section{Proposed Statistical Framework}
\label{section:statistical_significance}

\noindent \textbf{Fairness Metric}: In this work, \emph{equal opportunity} \cite{HardtEqualOpportunity2016} is used as the fairness criterion. A binary predictor $\hat{Y}$, which outputs an outcome $Y \in \{0,1\}$, satisfies equal opportunity with respect to protected attributed $A$ if:

\begin{equation}
\label{eqn:equalopp}
    P(\hat{Y} = 1|A = 0, Y = 1) = P(\hat{Y} = 1|A = 1, Y = 1), 
\end{equation}

\noindent where $Y = 1$ is typically considered as the ``positive'' or ``advantaged'' outcome and $A \in \{0, 1\}$ denotes group membership based on a protected attribute such as race (e.g., black {vs.} white) or gender (male {vs.} female). Equal opportunity implies that the \emph{true positive rates} for the two demographic groups under consideration are equal. 

\subsection{Framework for Verification}
\label{bootstrap_fairness_test}

A biometric system operating in the verification mode can be considered as a binary predictor $\hat{Y}$, whose outcome $Y \in \{0,1\}$ is based on matching two biometric samples. An outcome $Y = 0$ represents a ``non-match'' decision, while $Y = 1$ represents a ``match'' decision. Following the definition in Eq. \ref{eqn:equalopp}, we say that a biometric verification system is fair (in terms of equal opportunity) if the true match rates are equal across different demographic groups. Note that the true match rate is the proportion of mated comparisons that are correctly recognized as a match, and is equal to $1-FNMR$, where $FNMR$ is the false non-match rate of the fingerprint system.

Two key challenges need to be addressed in order to test the fairness of a system based on the criterion defined in Eq. \ref{eqn:equalopp}: (i) How to estimate the true positive rates accurately? and (ii) How to test for equality? Most of the existing works on fairness in the biometrics literature fail to address these challenges adequately. Note that $\hat{Y} = 1|A = 0, Y = 1$ and $\hat{Y} = 1|A = 1, Y = 1$  are Bernoulli random variables with parameters $p_0$ and $p_1$, respectively. In the case of biometric verification, maximum likelihood (point) estimates of the parameters $p_0$ and $p_1$ can be obtained by performing $n_0$ and $n_1$ mated comparisons for the two demographic groups ($A=0$ and $A=1$, respectively) and computing the corresponding true match rates. Let $\hat{p}_0$ and $\hat{p}_1$ be the MLE estimates of the parameters based on the given matcher on the given dataset. While these estimates can be compared directly to test for fairness, this approach fails to consider the uncertainty in the parameter estimation and lacks a solid theoretical foundation. For example, if $\hat{p}_0$ and $\hat{p}_1$ differ by a small amount, we cannot claim the two distributions are significantly different. 

Since the goal is to test the null hypothesis $H_0: p_0 = p_1$ versus the alternate hypothesis $H_A: p_0 \neq p_1$, the following test statistic could be used.

\begin{equation}
\label{eqn:parametricfairness}
    Z_p = \frac{\hat{p}_0-\hat{p}_1}{\sqrt{\hat{p}(1-\hat{p})\left(\frac{1}{n_0}+\frac{1}{n_1}\right)}},
\end{equation}

\noindent where $\hat{p} = \frac{n_0\hat{p}_0+n_1\hat{p}_1}{n_0+n_1}$ is the estimated true match rate for both the groups combined. The test statistic in eq. \ref{eqn:parametricfairness} asymptotically follows a standard normal distribution. Hence, for a $(1-\alpha)$ level test, the null hypothesis should be rejected if $|Z_p| > z_{1-\alpha/2}$, where $z_{\gamma}$ is the $\gamma$-quantile value of the standard normal distribution.

\noindent \textbf{Welch t-test}: The above hypothesis test relies on the assumption that the dataset used to estimate the true match rates is a random sample of the population. Since there may be biases in data collection and the mated comparisons may not be truly independent, an alternative is to evaluate the uncertainty in the estimation of $p_0$ and $p_1$ using the \emph{bootstrapping} (sampling with replacement) approach \cite{efron1992bootstrap}. Specifically, we draw $m_0$ and $m_1$ bootstrap samples from the available dataset to estimate $p_0$ and $p_1$, respectively. Let $\bar{p}_0$ ($\bar{p}_1$) be the mean true match rate for demographic group $A = 0$ ($A = 1$) based on the $m_0$ ($m_1$) bootstrap samples. Furthermore, let $s_{p_1}$ and $s_{p_2}$ be the standard deviation of the true match rates for the two groups. To evaluate statistical significance of the difference between the two means $\bar{p}_0$  and $\bar{p}_1$, we can use the two-sampled Welch's t-test \cite{welch}. The Welch t-test statistic is:

\begin{equation} 
\label{eqn:bootstrapfairness_eqvar}
Z_{w} = \frac{\bar{p}_0 - \bar{p}_1}{\sqrt{\frac{s_{p_0}^2}{m_0}+\frac{s_{p_1}^2}{m_1}}},
\end{equation}

\noindent and the null hypothesis $H_0: p_0 = p_1$ is rejected if $|Z_{w}| > t_{1-\alpha/2},\nu$, where $t_{\gamma,\nu}$ is the $\gamma$-quantile value of the t distribution with $\nu$ degrees of freedom. In this case, the degrees of freedom is given by:

\begin{equation} 
\nu = \frac{\left(\frac{s_{p_0}^2}{m_0} + \frac{s_{p_1}^2}{m_1}\right)^2}{\frac{\left(s_{p_0}^2/m_0\right)^2}{m_0-1} + \frac{\left(s_{p_1}^2/m_1\right)^2}{m_1-1}}.
\end{equation}

\noindent We refer to this test procedure as the \emph{bootstrap} fairness test.

\noindent \textbf{ANOVA test}: The Welch t-test is restricted to comparing two means. Statistical significance between the means of more than two groups can be evaluated using one way analysis of variance (ANOVA). Suppose that there are $k$ demographic groups and $p_1, p_2, \cdots,p_k$ are the true match rates for the $k$ groups. The null hypothesis is that all the $k$ groups have the same true match rate, i.e., $H_0: p_1 = p_2 = ... = p_k$. Let $p_{ij}$ be the true match rate computed from the $i^{th}$ bootstrap sample of group $j$, $i=1,2,\cdots,m$ and $j=1,2,\cdots,k$. Also, let $\bar{p}_j$ be the mean true match rate for the $m$ bootstrap samples of group $j$ and $\bar{\bar{p}}$ be the overall true match rate of the entire dataset. The test statistic \cite{Ross2017, STHLE1989259} is given by:

\begin{equation}
    F_0 = \frac{\frac{m}{(k-1)}\sum_{j=1}^{k}\left(\bar{p}_j-\bar{\bar{p}}\right)^2}{\frac{1}{k(m-1)}\sum_{i=1}^{m}\sum_{j=1}^{k}\left(p_{ij}-\bar{p}_j\right)^2}.
\end{equation}

One of the underlying assumptions for this test is that the variances of the sample populations are equal. The null hypothesis is rejected if the resulting $F_{0}$ statistic is greater than $F(\nu_1, \nu_2)_{critical}$, where $\nu_1$ and $\nu_2$ are the inter- and intra-class degrees of freedom. 

\begin{figure*}
\begin{center}
\subfloat[Verifinger 12.3 SDK]{\label{fig:veri_roc}\includegraphics[scale=0.38]{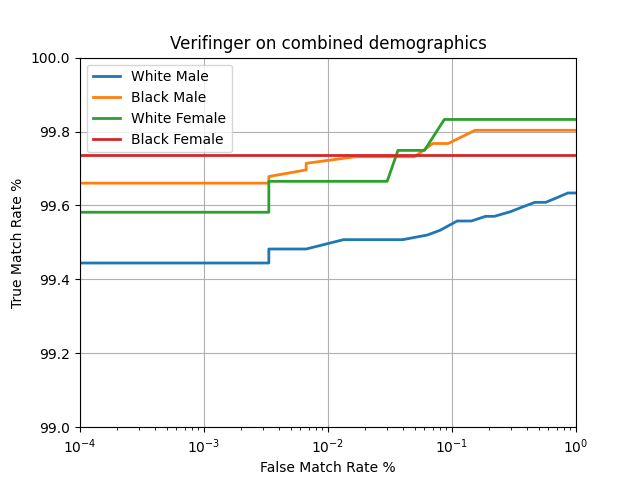}}
\subfloat[DeepPrint]{\label{fig:dp_roc}\includegraphics[scale=0.38]{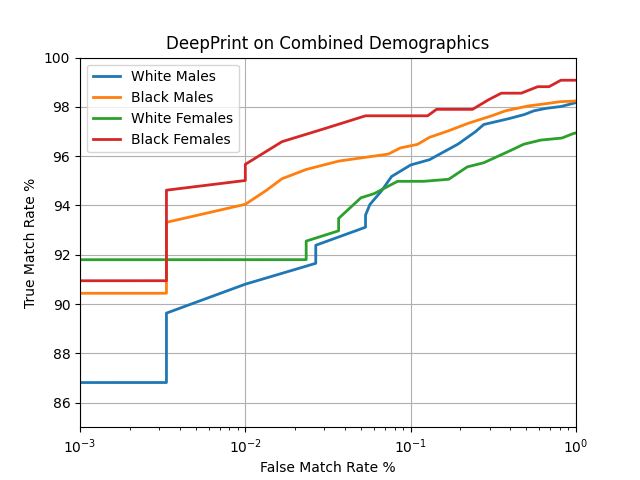}}
\captionsetup{justification=centering}
\caption{ROC Curves across the demographic categories combining race and gender for the two fingerprint matchers of database $D_1$. Curves are plotted using one of the 10 bootstrap samples generated and hence may not coincide with the $TMR$ \% reported in table \ref{tab:matcher_performance}.}
\label{fig:roc}
\end{center}
\vspace{-1.5em}
\end{figure*}

\begin{figure}
    \centering
    \includegraphics[scale=0.3]{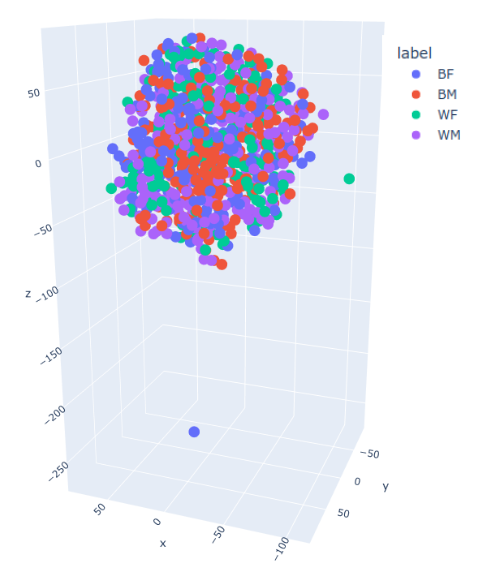}
    \captionsetup{justification=centering}
    \caption{3D t-SNE plot of $192$-dimensional embeddings generated from DeepPrint for database $D_1$. 200 samples were randomly drawn from each demographic group to provide better visual clarity. Note the two visible outliers that contribute towards the demographic differential observed in verification performance. These outliers do not constitute the entire set of outliers in dataset $D_1$.}
    \label{fig:tsne}
    \vspace{-1.5em}
\end{figure}

\subsection{Framework for Identification}

In the case of a fingerprint system operating in the closed set identification mode, a ``positive'' outcome ($Y = 1$) can be defined as the scenario where a query user is correctly identified among $N$ subjects in the gallery. Correspondingly, a ``negative'' outcome ($Y = 0$) implies that the query user is incorrectly matched against the wrong identity. Thus, $P(\hat{Y}=1|A=0,Y=1)$ and $P(\hat{Y}=1|A=1,Y=1)$ are the \emph{true positive identification rates} (TPIR) for the two demographic groups. Note that TPIR refers to the fraction of mated searches where the enrolled mate is within the top $R$ ranks, and is equal to $1-FNIR$, where $FNIR$ is the false negative identification rate \cite{maltoni2009handbook}. Thus, the same fairness testing framework derived for the verification mode can be applied for closed set identification, with $p_0$ and $p_1$ being the TPIR of the two demographic groups. The only modification is to account for the fact that TPIR is a function of both rank $R$ and the gallery size $N$.

Finally, for fingerprint systems operating in the \emph{open set identification mode}, the user will be greatly disadvantaged only when he is incorrectly assigned a wrong identity in the case of a mated search (query user is enrolled in the gallery) or is incorrectly identified as someone present in the gallery in the case of a non-mated search. Therefore, the performance metric that is critical from users' perspective is false positive identification rate (FPIR) and the demographic differential should be measured based on FPIR. To evaluate fairness, a common threshold is fixed for all the demographic categories and the discrepancies between $FPIR$ are considered for fairness evaluation, aligning the identification experiment with the protocol outlined in the NIST FRVT \cite{grother2019face, grother2022face}.

\section{Results}

\subsection{Databases and Recognition Systems}
\label{database}

The longitudinal forensics database $D_1$ consists of 15,468 subjects (unique fingers), each with a set of ten fingerprints taken each time the subject was arrested (on an average, eight times over a 12-year period). The second database $D_2$ contains 1,014 subjects collected using a CrossMatch Guardian 200 reader. Additional details about the number of subjects per demographic category can be found in Table \ref{tab:database}. These datasets were used to evaluate the fairness of a COTS fingerprint matcher Verifinger 12.3 SDK, along with a SOTA deep learning based matcher DeepPrint \cite{engelsma2019learning}. For verification, the COTS matcher provides a predefined threshold of 48 at a $FMR$ of 0.01\%. For DeepPrint, 0.83 was the threshold used - computed on NIST SD4 \cite{sd4} to obtain a $FMR$ of 0.01\%.

\subsection{Evaluation Protocol}

Verification and open-set identification performances are reported for the two matchers. To the best of our knowledge, most fingerprint recognition systems operate in one of these two modes of operation. Hence, we do not report results for closed-set identification. Based on the statistical framework developed in Section \ref{section:statistical_significance}, we claim that the demographic differential (bias) in a verification system is statistically significant if the null hypothesis $H_0:p_0=p_1$ is rejected. For verification, 10 bootstrapped samples are created from the entire population for each database $D_1$ and $D_2$. Using these 10 samples, both matchers are evaluated to obtain a mean $TMR$ $\bar{p}_j$  along with the standard deviation $s_j$, $j \in \{BF, BM, WF, WM\}$. Table \ref{tab:matcher_performance} summarizes the results and figure \ref{fig:roc} shows the associated ROC curves.

\subsection{Verification Bias}
\label{section:gender_bias}

\begin{table*}[ht]
    \centering
    \captionsetup{justification=centering}
    \caption{Mean $TMR$(\%) and standard deviation across all the demographic groups \tnote{1}.}
    \begin{threeparttable}

        \begin{tabular}{|c | c | c | c | c | c |} 
             \hline
             \textbf{Demographic Group} & \multicolumn{2}{c|}{\textbf{Dataset $D_1$}} & \multicolumn{2}{c|}{\textbf{Dataset $D_2$}} & avg. \# of genuine comparisons\\
             \noalign{\hrule height 1.2pt}
             & \textbf{Verifinger 12.3\tnote{2}} & \textbf{DeepPrint\tnote{3}} & \textbf{Verifinger 12.3\tnote{2}} & \textbf{DeepPrint\tnote{3}} & \\ [1.0ex]
             \hline
             {\includegraphics[scale=0.15]{images/bf.png}} BF       & 99.66 (\textpm0.23)    & 95.3 (\textpm0.49) & 100 (\textpm0)  & 91.07 (\textpm2.75) & 760\\ 
             {\includegraphics[scale=0.15]{images/bm.png}} BM         & 99.68 (\textpm0.07)  & 94.35 (\textpm0.39) & 100 (\textpm0) & 97.30 (\textpm2.11) & 5,619\\
             {\includegraphics[scale=0.15]{images/wf.png}} WF       & 99.53 (\textpm0.17)  & 92.03 (\textpm0.6)  & 100 (\textpm0) & 93.38 (\textpm1.3) & 1,188\\
             {\includegraphics[scale=0.15]{images/wm.png}} WM         & 99.46 (\textpm0.08)  & 91.18 (\textpm0.32) & 99.52 (\textpm0.51) & 94.33 (\textpm1.61) & 7,902\\
             
             \noalign{\hrule height 1.3pt}
             {\includegraphics[scale=0.15]{images/bm.png}} B         & 99.68 (\textpm0.08)  & 94.47 (\textpm0.38) & 100 (\textpm0) & 94.19 (\textpm3.96) & 6,379\\
             
             {\includegraphics[scale=0.15]{images/wf.png}} W       & 99.47 (\textpm0.08)    & 91.29 (\textpm0.28) & 99.75 (\textpm0.43) & 93.85 (\textpm1.54) & 9,089\\
             
             \noalign{\hrule height 1.3pt}
             {\includegraphics[scale=0.15]{images/bf.png}} F         & 99.58 (\textpm0.14)  & 93.3 (\textpm0.51) & 100 (\textpm0) & 92.23 (\textpm2.44) & 1,947\\
             {\includegraphics[scale=0.15]{images/wm.png}} M       & 99.55 (\textpm0.05)  & 92.5 (\textpm0.28)  & 99.76 (\textpm0.43) & 95.81 (\textpm2.4) & 13,521\\
             \hline
             
             \hline
        \end{tabular}
    \begin{tablenotes}
    \item[1]{Mean $TMR$ \% and standard deviation were obtained using the 10 bootstrap samples.}
    \item[2]{Threshold = 48. Specified by vendor.}
    \item[3]{Threshold = 0.83. Computed based on matcher performance on NIST SD4 \cite{sd4} at a $FMR = 0.01\%$ .}
    \end{tablenotes}
    \end{threeparttable}
    \label{tab:matcher_performance}
\end{table*}

\par The mean $TMR$ values for dataset $D_1$ in Table \ref{tab:matcher_performance} indicate that Verifinger has a higher performance on BM compared to BF and a lower performance on WM compared to WF. Overall, this matcher has a higher $TMR$ on black subjects compared to white subjects and almost equal $TMR$ for male and female subjects. DeepPrint also shows a higher performance on black subjects compared to white subjects and a higher performance on females compared to males. To evaluate statistical significance, we test $H_0: p_0 = p_1$ versus $H_A: p_0 \neq p_1$ where $p_0$ and $p_1$ are the mean $TMR$ (\%) of the two groups being evaluated. Table \ref{tab:final_significance} consolidates the pairwise Welch's t-test results for all the relevant comparisons. Our analysis in this paper is largely focused on dataset $D_1$ because of its large size. Verifinger does exhibit significant difference in $TMR$ between WM and BM groups but the biases are more prevalent in DeepPrint with the differentials in $TMR$ being statistically significant for each pairwise comparison when evaluated on $D_1$. ANOVA test on the mean $TMR$ from Verifinger for BF, BM, WF and WM yields $F_{0}=4.56$ with inter-class and intra-class degrees of freedom $\nu_1=3$ and $\nu_2=36$. Since $F_{0} > F(3,36)_{critical}=2.87$, we can conclude that there \textbf{is} demographic differential between the $TMR$ of the 4 groups when evaluated using Verifinger. Similarly, there is also a bias present in DeepPrint because $F_{0}=173.46>F(3,36)_{critical}=2.87$. This is not surprising given the existence of pairwise biases.

\begin{figure}
    \centering
    \includegraphics[scale=0.3]{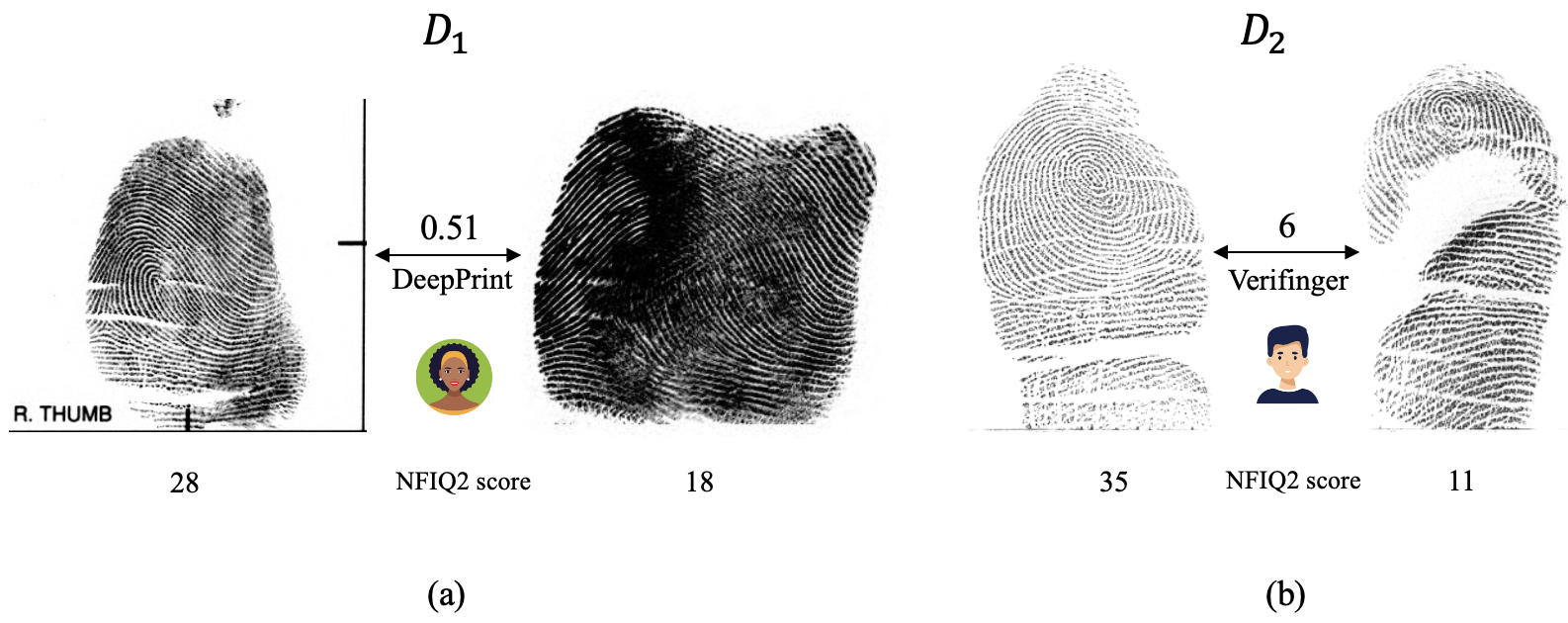}
    \captionsetup{justification=centering}
    \caption{Examples of genuine comparisons below the preset threshold. One example is shown for both databases $D_1$ and $D_2$ along with the NFIQ2 scores below each image and an avatar showing the demographic group. Additionally, the similarity score from a matcher is also shown. Note that the images on the right in (a) and (b) have a significantly lower NFIQ2 score compared to their counterparts on the left leading to low match scores.} 
    \label{fig:misclassified}
    \vspace{-1em}
\end{figure}

\vspace{-1.0em}

\par Verifinger and DeepPrint are different types of matchers - the former is minutiae-based and the latter is based on a deep neural network. Since DeepPrint has lower overall accuracy and reliability compared to Verifinger, it is more prone to demographic bias due to the larger range of $TMR$ values. Since there are an average of only 760 genuine comparisons for BF, even a few noisy similarity scores can lead to statistically significant differentials among the demographic groups. Figure \ref{fig:tsne} shows a 3-D t-SNE plot of the embeddings obtained from DeepPrint for dataset $D_1$. It shows embeddings for 200 randomly selected subjects for each demographic group. We can see two clear outliers in the figure - one from BF and one from WF. It should be noted that these outliers are not the only ones present in the dataset but are the ones that were included in the random sampling. Figure \ref{fig:misclassified} shows examples of genuine similarity scores below the matcher threshold. In Fig. 4 (a), the image on the right is blotted and the ridge patterns are not clearly visible. In Fig. 4 (b), we see that the image on the right has a large gap in the middle, which explains the low match score of 6. Thus, the $TMR$ differentials between the demographic groups can be explained by such outliers to a large degree.

\begin{figure}
    \centering
    \includegraphics[scale=0.45]{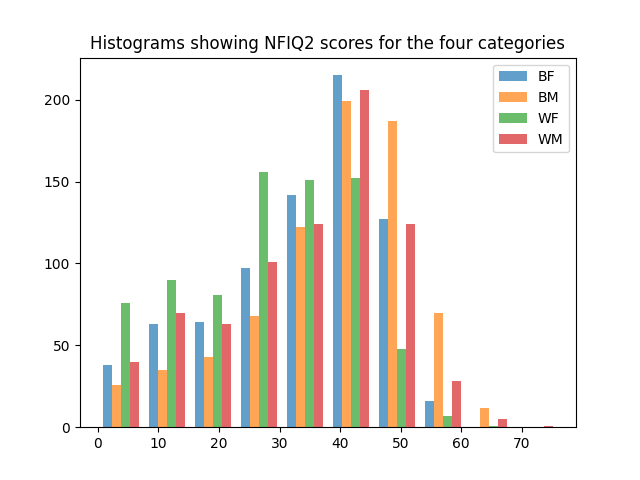}
    \captionsetup{justification=centering}
    \caption{Distribution of NFIQ2 scores for the four demographic groups from dataset $D_1$. The NFIQ2 scores (x-axis) range from $[0,100]$ with a higher score corresponding to a higher image quality. In this figure, the same number of images were sampled for each demographic category.}
    \label{fig:nfiq_stats}
\end{figure}

\par Additionally, Figure \ref{fig:nfiq_stats} shows that female subjects have a lower mean NFIQ2 score compared to male subjects. Intuitively, the matchers will perform better on the images with higher quality - assuming a correlation between image quality and matcher performance. However, that is not the case for DeepPrint - it performs better on female subjects compared to male. This outcome mirrors the observations made by Marasco \cite{9186012} and leads to the conclusion that the NFIQ2 quality metric may itself be biased.

\begin{table*}[ht]
    \centering
    \captionsetup{justification=centering}
    \caption{Pairwise t-test results for \textbf{verification} on database $D_1$ ($\alpha = 0.05$). For both matchers, the absolute value of the t-statistic ($|Z_{w}|$), the degrees of freedom ($\nu$), and the hypothesis testing result are provided. Here,  \emph{Yes} means that the null hypothesis was rejected (the observed demographic differential is statistically significant) and \emph{No} means the null hypothesis cannot be rejected.}
    \begin{threeparttable}
    \begin{tabular}{|c || c | c | c | c | c | c |} 
         \hline
         Demographic Group & \multicolumn{3}{c|}{\textbf{Verifinger 12.3}} & \multicolumn{3}{c|}{\textbf{DeepPrint}} \\ [1.0ex]
         \noalign{\hrule height 1.2pt}
         & \(\displaystyle |Z_{w}|\) & \(\displaystyle \nu \) &  \(\displaystyle H_0 \) rejected? & \(\displaystyle |Z_{w}|\) &  \(\displaystyle \nu \) & \(\displaystyle H_0 \) rejected? \\
         \hline
         WF and WM             &  1.18  & 12.8 & No & 3.95 & 13.73 & \textbf{Yes}   \\
         \hline
         BF and BM            &  0.26 & 10.65 & No & 4.79 & 17.14 & \textbf{Yes}  \\
         \hline
         WM and BM              & 6.54  & 17.69 & \textbf{Yes} & 19.87 & 17.34 & \textbf{Yes}  \\
         \hline
         WF and BF            & 1.44 & 16.57 & No & 13.34 & 17.31 & \textbf{Yes}   \\
         \noalign{\hrule height 1.2pt}
         
         F and M            & 0.64  & 11.25 & No & 4.34 & 13.97 & \textbf{Yes}   \\
         \noalign{\hrule height 1.2pt}
         
         B and W          & 5.86 & 18 & \textbf{Yes} & 21.3 & 16.55 & \textbf{Yes}   \\
         \noalign{\hrule height 1.2pt}
    \end{tabular}
    \end{threeparttable}
    \label{tab:final_significance}
\end{table*}

\begin{table*}[ht]
    \centering
    \captionsetup{justification=centering}
    \caption{Pairwise t-test results for \textbf{open set identification} on database $D_1$ ($\alpha = 0.05$).}
    \begin{threeparttable}
    \begin{tabular}{|c || c | c | c | c | c | c |} 
         \hline
         Demographic Group & \multicolumn{3}{c|}{\textbf{Verifinger 12.3}} & \multicolumn{3}{c|}{\textbf{DeepPrint}} \\ [1.0ex]
         \noalign{\hrule height 1.2pt}
         & \(\displaystyle |Z_{w}|\) & \(\displaystyle \nu \) &  \(\displaystyle H_0 \) rejected? & \(\displaystyle |Z_{w}|\) &  \(\displaystyle \nu \) & \(\displaystyle H_0 \) rejected? \\
         \hline
         WF and WM             &  5.2  & 7.60 & \textbf{Yes} & 3.0 & 6.71 & \textbf{Yes}   \\
         \hline
         BF and BM            & 2.34  & 4.07 & No & 1.50 & 7.93 & No  \\
         \hline
         WM and BM              & 1.93  & 7.83 & No & 2.77 & 7.34 & \textbf{Yes}  \\
         \hline
         WF and BF            & 7.02 & 4.03 & \textbf{Yes} & 4.39 & 5.35 & \textbf{Yes}   \\
         \noalign{\hrule height 1.2pt}
    \end{tabular}
    \end{threeparttable}
    \label{tab:identification_significance}
\end{table*}

\subsection{Open-set Identification Bias}
\label{search_bias}

\par For open-set identification, we follow the experiment design described in \cite{tabassi2014performance} and report the performance based on the two metrics defined in the NIST FRVT report, $FPIR$ and $FNIR$ \cite{grother2019face, grother2022face}. FRVT defines the $FPIR$ as the fraction of non-mated searches that returned one or more similarity scores above the specified threshold and $FNIR$ as the fraction of mated searches where the mate has a similarity score less than the threshold or was found outside $R$ ranks \cite{grother2019face}. In this case $R$ was selected to be 5. 

\begin{figure}
    \centering
    \includegraphics[scale=0.4]{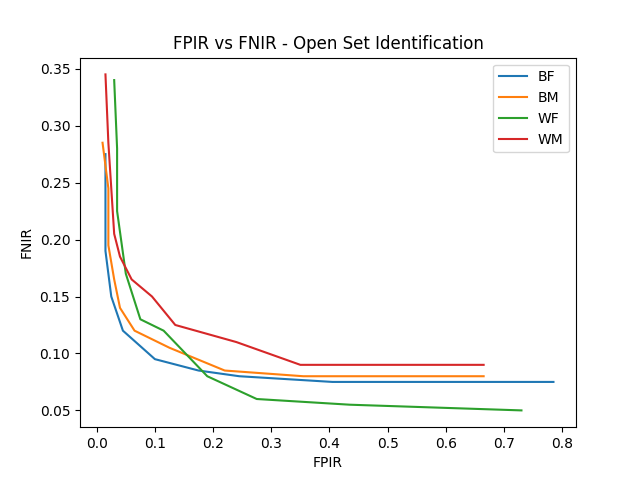}
    \captionsetup{justification=centering}
    \caption{FPIR vs FNIR for DeepPrint on database $D_1$. Matcher thresholds are varied between [0.5, 0.95].}
    \label{fig:open_set}
\end{figure}

\par To evaluate open-set identification accuracy, a gallery is constructed consisting of 100,000 unique fingers\footnote{No demographic information is available for these 100,000 images and there are no overlapping identities.} from a separate database as distractors along with impressions from three demographic groups (from $D_1$) selected as follows; if we are evaluating performance on BF, the three groups in the gallery will be BM, WF, WM. Out of the 4 groups, black females have the least number of probe subjects (762). To maintain consistency, the same number of probe subjects were sampled from all four groups. As described by Tabassi  \cite{tabassi2014performance}, the probe cohort is split up into two categories; subjects that have a mate in the gallery ($P_G$) and subjects that do not have a mate ($P_N$)\footnote{$P_N$ and $P_G$ are disjoint from each other.}. The number of subjects with a mate in the gallery is chosen to be 200.  This brings the total gallery size to 102,486 identities \footnote{$100$k unique fingers $+$ ($3 \times 762$) subjects from the 3 other demographic categories $+$ $200$ mates of $P_G$}, in alignment with the procedure specified in \cite{jain2011handbook} that recommends one biometric sample per identity in the gallery.

\begin{table}[]
    \centering
    \captionsetup{justification=centering}
    \caption{FPIR \% for Verifinger and DeepPrint at fixed thresholds for database $D_1$.}
    \vspace{2pt}
    \begin{threeparttable}

        \begin{tabular}{|c|c|c|c| } 
             \hline
             \textbf{Demographic Group}  & \textbf{Verifinger} & \textbf{DeepPrint}\\ [1.0ex]
             \hline
             BF  & 8.04 (\textpm0.11) &  2.35 (\textpm0.5)\\  
             \hline
             BM  & 6.8 (\textpm1.17) & 1.85 (\textpm0.55)\\  
             \hline
             WF  & 13.45 (\textpm1.72) & 4.9 (\textpm1.2)\\  
             \hline
             WM  & 8.35 (\textpm1.36) & 3.0 (\textpm0.75)\\  
             \hline
        \end{tabular}
        \end{threeparttable}
    \label{tab:open_set}
    \vspace{-1.5em}
\end{table}

\par Figure \ref{fig:open_set} and Table \ref{tab:open_set} show the open-set identification performance in terms of $FPIR$ and $FNIR$ for DeepPrint. To ensure that the $FPIR$ values are comparable, the threshold is fixed across all four demographic groups. In the case of Verifinger, a threshold of 48 is provided by the vendor. For DeepPrint, a matcher threshold is computed to obtain a desired $FNIR$ (0.1) on a specific demographic group (BF) and that same threshold (0.8904) is applied to all four demographic groups. From Tables \ref{tab:open_set} and \ref{tab:identification_significance}, it is observed that both Verifinger and DeepPrint exhibit the lowest performance on WF (highest mean $FPIR$ at the given threshold) and the highest performance on BM (lowest mean $FPIR$). The differentials observed in Verifinger are statistically significant when the WF demographic group is included in the comparison. DeepPrint showed statistically significant differentials in all cases except when comparing BF and BM.

\section{Discussion and Summary}

\par Two clear trends emerge from this study: (i) in the case of verification, more accurate fingerprint matchers are less likely to exhibit bias (demographic differentials), and (ii) most observed demographic differentials can be explained by the poor quality of some fingerprint images, which leads to outliers in the genuine score distribution. Depending on database size, even a small number of noisy mated comparisons can cause a significant drop in the performance of the matcher, thereby providing false evidence of bias. For example, out of the 9,112 total white subjects (and genuine similarity scores), if only 13 (0.15\%) of the genuine similarity scores below the threshold were increased to be above the threshold, the bias between black and white subjects in Verifinger would disappear. This would also eliminate the bias between WM and BM. Therefore, our study strongly indicates that the differentials in fingerprint recognition performance between demographic groups depend on external factors that are unrelated to the demographic of the user. Hence, it cannot be said with confidence that the biases encountered in facial recognition are also present in fingerprint recognition.

{\small
\bibliography{cite}
\bibliographystyle{ieeetr}
}

\end{document}